\def\year{2020}\relax
\documentclass[letterpaper]{article} % DO NOT CHANGE THIS
\usepackage{aaai20}  % DO NOT CHANGE THIS
\usepackage{times}  % DO NOT CHANGE THIS
\usepackage{helvet} % DO NOT CHANGE THIS
\usepackage{courier}  % DO NOT CHANGE THIS
\usepackage[hyphens]{url}  % DO NOT CHANGE THIS
\usepackage{graphicx} % DO NOT CHANGE THIS
\usepackage{graphicx}  % DO NOT CHANGE THIS
\usepackage{amsfonts}
\usepackage{amssymb}
\usepackage{xspace}
\usepackage{booktabs}
\usepackage[ruled,vlined]{algorithm2e}
\usepackage{amsmath,bm}
\usepackage{dsfont}
\usepackage{multirow}
\usepackage [autostyle, english = american]{csquotes}
\usepackage{xcolor}
\usepackage{subcaption} 
\MakeOuterQuote{"}
\usepackage{ntheorem,lipsum} %remove italic font in definition
\theorembodyfont{\upshape} %remove italic font in definition
\newtheorem{definition}{Definition}

\newcommand{\mname}{\texttt{CASTER}\xspace}
\usepackage[shortlabels]{enumitem}
\begin{document}
\title{\mname: Predicting Drug Interactions with \\ Chemical Substructure Representation}

\author{
\textbf{Kexin Huang,\textsuperscript{\rm 1,2}
Cao Xiao,\textsuperscript{\rm 2}
Trong Nghia Hoang,\textsuperscript{\rm 3}
Lucas M. Glass,\textsuperscript{\rm 2}
Jimeng Sun\textsuperscript{\rm 4}} \\
\textsuperscript{\rm 1} Harvard T. H. Chan School of Public Health, Harvard University, Boston, MA, USA \\
\textsuperscript{\rm 2} Analytic Center of Excellence, IQVIA, Cambridge, MA, USA \\ 
\textsuperscript{\rm 3} MIT-IBM Watson AI Lab, IBM Research, Cambridge, MA, USA \\
\textsuperscript{\rm 4}  Computational Science and Engineering, Georgia Institute of Technology, Atlanta, GA, USA\\
kexinhuang@hsph.harvard.edu, cao.xiao@iqvia.com, nghiaht@ibm.com, lucas.glass@iqvia.com, jsun@cc.gatech.edu
}
\maketitle
\begin{abstract}
Adverse drug-drug interactions (DDIs) remain a leading cause of morbidity and mortality. Identifying potential DDIs during the drug design process is critical for patients and society. Although several computational models have been proposed for DDI prediction, there are still limitations: (1) specialized design of drug representation for DDI predictions is lacking; (2) predictions are based on limited labelled data and do not generalize well to unseen drugs or DDIs; and (3) models are characterized by a large number of parameters, thus are hard to interpret. In this work, we develop a \texttt{C}hemic\texttt{A}l \texttt{S}ubstruc\texttt{T}ur\texttt{E} \texttt{R}epresentation (\mname) framework that predicts DDIs given chemical structures of drugs. \mname aims to mitigate these limitations via
(1) a sequential pattern mining module rooted in the DDI mechanism to efficiently characterize  functional sub-structures of drugs; (2) an auto-encoding module that leverages both labelled and unlabelled chemical structure data to improve predictive accuracy and generalizability; and (3) a dictionary learning module that explains the prediction via a small set of coefficients which measure the relevance of each input sub-structures to the DDI outcome. We evaluated \mname on two real-world DDI datasets and showed that it performed better than state-of-the-art baselines and provided interpretable predictions.
\end{abstract}

\section{Introduction}
\label{intro}
Adverse drug-drug interactions (DDIs) are caused by pharmacological interactions of drugs. They result in a large number of morbidity and mortality, and incur huge medical costs~\cite{Giacomini2007,onakpoya2016post}. Thus gaining accurate and comprehensive knowledge of DDIs, especially during the drug design process, is important to both patients and pharmaceutical industry.
Traditional strategies of gaining DDI knowledge includes preclinical \emph{in vitro} safety profiling and clinical safety trials, however they are restricted in terms of small scales, long duration, and huge costs~\cite{whitehead}. Recently, deep learning (DL) models that leverage massive biomedical data for efficient DDI predictions emerged as a promising direction~\cite{lo2018machine,ryu2018deep,Xiao:2017:ADR:3298239.3298470,AAAI1714292,core}. These methods are based on the assumption that drugs with similar representations (of chemical structure) will have similar properties (e.g., DDIs). Despite the reported good performance, these works still have the following limitations:
\begin{enumerate}[leftmargin=*, nosep]
    \item \textbf{Lack of specialized drug representation for DDI prediction}. One of the major mechanism of drug interactions results from the chemical reactions among only a few functional sub-structures of the entire drug's molecular structure ~\cite{silverman2014organic}, while the remaining substructures are less relevant. Many drug pairs that are not similar in terms of DDI interaction can still have significant overlap on irrelevant substructures. Previous works \cite{ryu2018deep,gomez2018automatic,jaeger2018mol2vec} often generate drug representations using the entire chemical representation, which causes the learned representations to be potentially biased toward irrelevant sub-structures. This undermines the learned drug similarity and DDI predictions.
    \item \textbf{Limited labels and generalizability}. Some of the previous methods need external biomedical knowledge for improved performance and cannot be generalized to drugs in early development phase~\cite{ma2018drug,Ferdousi17,zhang2015label}. Others rely on a small set of labelled training data, which impairs their generalizability to new drugs or DDIs~\cite{ryu2018deep,zhang2015label}. 
    \item \textbf{Non-interpretable prediction.} Although DL models show good performance in DDI prediction, they often produce predictions that are characterized by a large number of parameters, which is hard to interpret~\cite{gomez2018automatic,jaeger2018mol2vec}.
\end{enumerate}
\begin{figure*}[t]
    \centering
    \includegraphics[width = 0.7\textwidth]{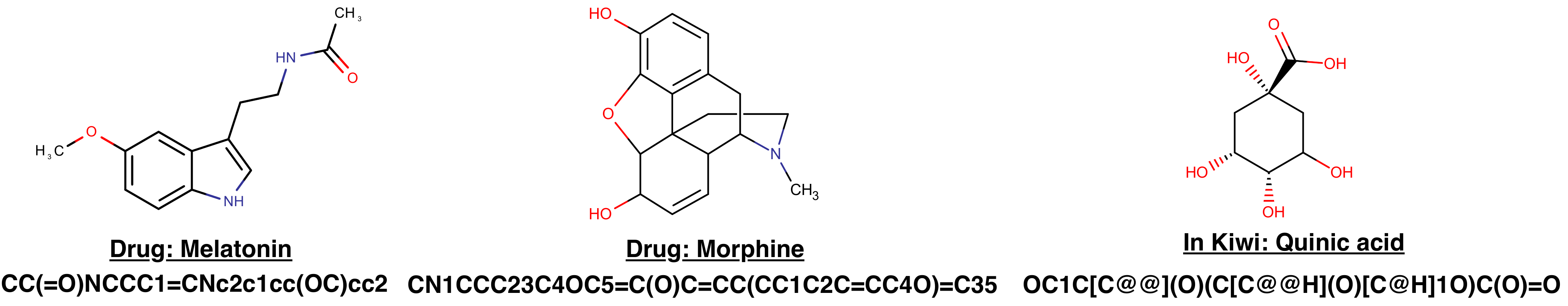}
    \caption{Examples of SMILES strings representing the molecular graphs of drugs and food constituents.}
    \label{smiles}
\end{figure*}
In this paper, we present a new \texttt{C}hemic\texttt{A}l \texttt{S}ubstruc\texttt{T}ur\texttt{E} \texttt{R}epresentation (\mname) framework that performs DDI predictions based on chemical structure data of drugs. \mname mitigates the aforementioned limitations via the following technical contributions: 

\begin{enumerate}[leftmargin=*, nosep]
    \item {\bf Specialized representation for DDI.} We developed a sequential pattern mining  module~(Alg.~\ref{cbpe}) to account for the interaction mechanism between drugs, which boils down to the reaction between their functional sub-structures (i.e., functional groups)~\cite{silverman2014organic}. This allows us to explicitly model a pair of drugs as a set of their functional sub-structures (Section~\ref{representation}). Unlike previous works, this representation can be learned using both labelled and unlabelled data (i.e., drug pairs that have no reported interaction) and is therefore more informative towards DDI prediction as demonstrated in our experiment (Section~\ref{experiment}). 
    \item {\bf Improved generalizability.} \mname only uses chemical structure data that any drug has as model input, which extends the learning beyond DDI to include any compound pairs with associated chemical information, such as drug-food interaction (DFI). Food compounds are described by the same set of chemical features, thus using unlabeled drug-food pair can help improve drug-drug representation's generalizability for better DDI performance. To achieve this, \mname has a deep auto-encoding module that extracts information from both labelled and unlabelled drug pairs to embed these patterns into a latent space (Section~\ref{embedding}). The embedded representation can be generalized to new drug pairs with reduced bias towards irrelevant or noisy artifacts in training data, thus improving generalizability.
    \item {\bf Interpretable prediction.} \mname has a dictionary learning module that generates a small set of coefficients to measure the relevance of each input sub-structures to the DDI outcomes, thus making it more interpretable\footnote{This is only to understand how \mname makes predictions. It is NOT our aim to interpret the prediction outcome in terms of the chemical process that regulates drug interactions.} to human practitioners than the predictions produced by existing DL methods. This is achieved by projecting drug pairs onto the subspace defined by the above generalizable embeddings of the frequent sub-structures (Section~\ref{prediction}).
\end{enumerate}
We compared \mname with several state-of-the-art baselines including DeepDDI~\cite{ryu2018deep}, mol2vec~\cite{jaeger2018mol2vec} and molecular VAE~\cite{gomez2018automatic} on two real-world DDI datasets. Experimental results show \mname outperformed the best baselines consistently (Section~\ref{experiment}). We also show by a case study that \mname produces interpretable results that capture meaningful sub-structures with respect to the DDI outcomes.

\section{Method}
\label{method}

\begin{figure*}[t]
    \centering
    \includegraphics[width =0.8\textwidth]{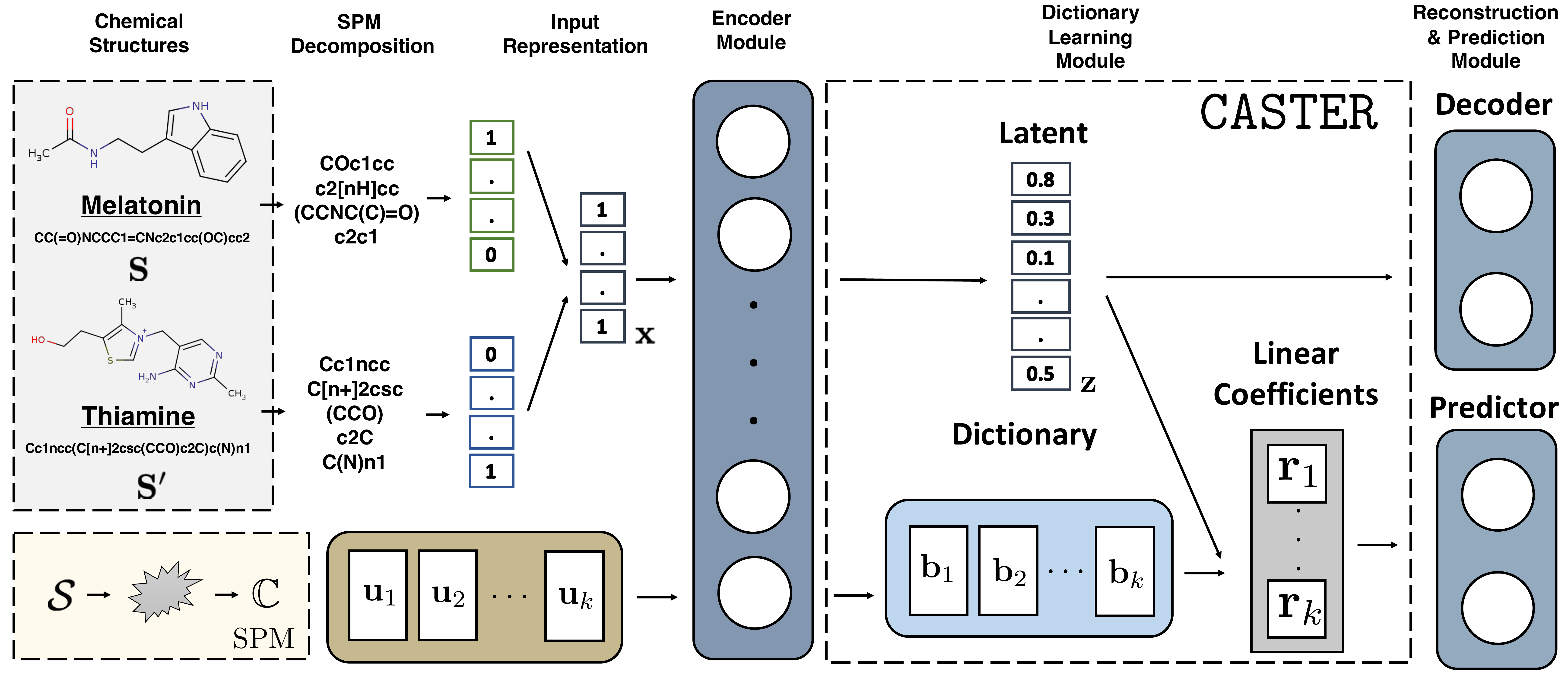}
    \caption{\mname workflow: (a) \mname extracts frequent substructures $\mathbb{C}$ from the molecular database $\mathcal{S}$ via SPM (see Alg.~\ref{cbpe}); (b)  $\mathbf{x}$ is generated for each input pair $(\mathbf{S}, \mathbf{S}')$; (c) the functional representation $\mathbf{x}$ is embedded into a latent space via minimizing a reconstruction loss, which results in a latent feature vector $\mathbf{z}$; the function representations of each frequent substructures $\{\mathbf{u}_i\}_{i=1}^k$ are also embedded into the latent space to yield a dictionary entry  $\{\mathbf{b}_i\}_{i=1}^k$ respectively; (d) the latent feature $\mathbf{z}$ is projected onto the subspace of $\{\mathbf{b}_i\}_{i=1}^k$ and results in linear coefficients $\{\mathbf{r}_i\}_{i=1}^k$; and (e) $\{\mathbf{r}_i\}_{i=1}^k$ are used as features for training DDI prediction module. All components are in one trainable pipeline which is optimized end-to-end by minimizing both the reconstruction and prediction losses.}
    \label{caster}
\end{figure*}

This section presents our \mname framework (Fig.~\ref{caster}). Our problem settings are summarized in Section~\ref{formulation}. Then, Section~\ref{representation} describes a functional representation of drugs which is inspired by the mechanism of drug interactions. Section~\ref{embedding} develops a deep auto-encoding method that leverages both labelled and unlabelled data to further embed the functional representation of drugs into a parsimonious latent space that can be generalized to new drugs. Section~\ref{prediction} then explains how this representation can be leveraged to generate accurate and interpretable predictions.

\subsection{Problem Settings}
\label{formulation}

The DDI prediction task is about developing a computational model that accepts two drugs (or their representations) as inputs and generate an output prediction indicating whether there exists an interaction between them. In this task, each drug can be encoded by the simplified molecular-input line-entry system (SMILES) \cite{smiles}. In particular, the associated SMILES string $\mathbf{S}$ for each drug is a sequence of symbols of the chemical atoms and bonds in its depth-first traversal order of its molecular structure graph (see Fig.~\ref{smiles}) where nodes are atoms and edges are bonds between them.

Our dataset consists of (a) a set of SMILES strings $\mathcal{S} = \{\mathbf{S}_1, \ldots, \mathbf{S}_n\}$ where $n$ is the total number of drugs, (b) a set $\mathcal{I} = \{(\mathbf{S}_t, \mathbf{S}_t')\}_{t=1}^{m}$  
of $m$ drug pairs reported to have interactions; and (c) a set $\mathcal{U} = (\mathcal{S}\times\mathcal{S})\setminus\mathcal{I}$ of length $\tau$ drug pairs that have not been reported with interactions.

To predict drug interactions, we need to learn a mapping $\mathcal{G}: \mathcal{S} \times \mathcal{S} \rightarrow [0, 1]$ from a drug-drug pair $(\mathbf{S}, \mathbf{S}') \in \mathcal{S} \times \mathcal{S}$ 
to a probability that indicates the chance that $\mathbf{S}$ and $\mathbf{S}'$ will have interaction. 

This will be achieved by learning an embedded functional representation $\mathbf{z} \in \mathbb{R}^d$ for each drug via a deep embedding module $\mathcal{E}: \mathcal{S} \rightarrow \mathbb{R}^d$ (Sections~\ref{representation} and~\ref{embedding}), and a mapping $\mathcal{G}: \mathcal{E} \times \mathcal{E} \rightarrow [0,1]$ from the embedded representation $(\mathbf{z},\mathbf{z}')$ of a drug pair to their interaction outcome (Section~\ref{prediction}).

\subsection{Generating Functional Representations}
\label{representation}

Instead of generating a wholistic representation for input chemical, we generate a functional representation using sub-structures information. This section tackles the frequent sequential pattern mining task to generate recurring chemical sub-structures in our database $\mathcal{S}$ of molecular representations of drugs. In particular, we define a frequent pattern formally as:

\begin{algorithm}[h]
\SetAlgoLined
Initialize $\mathbb{V}$ to the set of all atoms and bonds, $\mathbb{W}$ as the set of tokenized SMILES strings input\\
Input $\eta$ as the practitioner-specified frequency threshold, and $\ell$ as the maximum size of $\mathbb{V}$ \\
\For{$t = 1 \ldots \ell$}{
    $\textsc{(A, B), freq} \leftarrow \mathrm{scan}\ \mathbb{W}$\\ $// \textsc{(A,B), freq}~\text{are the frequentest pair and its frequency}$ \\
    \If{$\textsc{freq} < \eta$}{
        $\mathrm{break}$ \hfill // \text{frequency lower than threshold}\\ 
    } 
    $\mathbb{W} \leftarrow \mathrm{find} \textsc{(A, B)} \in \mathbb{W},  \mathrm{replace}~\mathrm{with}\  \textsc{(AB)}$ \\ // \text{update $\mathbb{W}$ with the new token \textsc{(AB)}}\\
    $\mathbb{V} \leftarrow \mathbb{V} \cup \textsc{(AB)}$ \\ 
    // \text{add \textsc{(AB)} to the vocabulary set $\mathbb{V}$}\\
}
\caption{The Chemical Sequential Pattern Mining Algorithm}
\label{cbpe}
\end{algorithm}

\begin{definition}[Frequent Chemical Substructures]
Let $\mathbf{S} \in \mathcal{S}$ be a SMILES string representing the molecular graph of a drug in depth-first traversal order (see Fig.~\ref{smiles}) and let $\mathbf{C}$ be a consecutive sub-string of it. Then, $\mathbf{C}$ also corresponds to a depth-first traversal representation of a molecular sub-graph. $\mathbf{C}$ is a frequent substructure if its occurring frequency in $\mathcal{S}$ is above a threshold $\eta$. 
\end{definition}

This can be achieved by observing that by definition sub-structures (or sub-graphs) that appear across different drugs will be represented by the same SMILES sub-strings thanks to their depth-first traversal representations. As such, frequent chemical sub-graphs can be extracted efficiently by finding frequent SMILES sub-strings across the entire dataset. This is similar to the concept of sub-word units in the natural language processing domain~\cite{gage1994new,sennrich2015neural}. More concretely, we propose a data-driven Chemical Sequential Pattern Mining algorithm (SPM) algorith~\cite{espf} see Algo.~\ref{cbpe}. Given an arbitrary input chemical SMILES string $\mathbf{S}$, SPM generates a set of discrete frequent sub-structures. This discreteness enables explainability that previous fingerprints are not capable of. The generated substructures will be used to induce functional representations for all drug pairs in both labelled $\mathcal{I}$ and unlabelled $\mathcal{U}$ datasets, which is defined formally in Definition 2 below. 

\begin{definition}[Functional Representation]
Let $\mathbb{C} = \{\mathbf{C}_1, \ldots, \mathbf{C}_k\}$ denote the set of frequent substructures. Each drug pair $(\mathbf{S},\mathbf{S}')$ is now represented via a $k$-dimensional multi-hot vector $\mathbf{x} = [\mathbf{x}^{(1)}\  \mathbf{x}^{(2)} \ldots \mathbf{x}^{(k)}]$ where $\mathbf{x}^{(i)} = 1$ if $\mathbf{C}_i \in \mathbf{S}$ and $\mathbf{C}_i \in \mathbf{S}'$, and $\mathbf{x}^{(i)} = 0$ otherwise. The resulting vector $\mathbf{x}$ is formally defined as the functional representation of $(\mathbf{S}, \mathbf{S}')$.
\end{definition}

Note that this functional representation maps drug pairs to the frequent substructures. Different substructures may interact with each other which lead to drug interactions. We find this representation alone has better predictive values than classic fixed-sized drug fingerprints~\cite{ryu2018deep,rogers2010extended,riniker2013open,james2004daylight} in DDI prediction task.

\subsection{Latent Feature Embedding}
\label{embedding}

This section develops a latent feature embedding module that extracts information from both labelled and unlabelled drug pairs to further embed the above functional representations into a latent space, which can be generalized to new drug pairs. This includes an encoder, a decoder and a reconstruction loss components which are detailed below. 

\subsubsection{\bf Encoder.} For each functional representation $\mathbf{x}_t \in \{0,1\}^k$ of a drug-drug or drug-food pair $(\mathbf{S}_t, \mathbf{S}'_t)$, a $d$-dimensional ($d \ll k$) latent feature embedding is generated via a neural network (NN) parameterized with weights $\mathbf{W}_{\mathrm{e}}$ and biases $\mathbf{b}_{\mathrm{e}}$:
\begin{equation}
\label{eq_encoder}
\mathbf{z}_t = \mathbf{W}_\mathrm{e}\mathbf{x}_t + \mathbf{b}_\mathrm{e} \ ,
\end{equation}
\subsubsection{Decoder.} To reconstruct the functional representation $\mathbf{x}_t$ given the latent embedding $\mathbf{z}_t$, we define another NN framework similarly parameterized by $(\mathbf{W}_\mathrm{d}, \mathbf{b}_\mathrm{d})$ that maps $\mathbf{z}_t$ back to 
\begin{equation}
\label{eq_decoder}
\widehat{\mathbf{x}}_t = \mathrm{sigmoid}\Big(\mathbf{W}_\mathrm{d}\mathbf{z}_t + \mathbf{b}_\mathrm{d}\Big) \ ,
\end{equation}
where $\mathrm{sigmoid}(\mathbf{a}) = [\sigma(a_1) \ldots \sigma(a_k)]$ is a point-wise function with $\sigma(a_i) = 1 / (1 + \mathrm{exp}(-a_i))$.

\subsubsection{Reconstruction Loss.} The encoding and decoding parameters $(\mathbf{W}_\mathrm{d}, \mathbf{b}_\mathrm{d})$ and $(\mathbf{W}_\mathrm{e}, \mathbf{b}_\mathrm{e})$ can be learned via minimizing the following reconstruction loss: 
\begin{equation}
\label{eq_lr}
\mathbf{L}_r(\mathbf{x}_t, \widehat{\mathbf{x}}_t) = \sum_{i=1}^{k} \left(\mathbf{x}_t^{(i)}\log\left(\widehat{\mathbf{x}}_t^{(i)}\right) \hspace{-1mm}+\hspace{-1mm} \left(1-\mathbf{x}_t^{(i)}\right)\log\left(1-\widehat{\mathbf{x}}_t^{(i)}\right)\right).
\end{equation}
Eq.~\eqref{eq_lr} succinctly summarizes our latent feature embedding procedure, which is completely unsupervised since optimizing Eq.~\eqref{eq_lr} only requires access to unlabelled drug pairs as training data. This allows \mname to tap into the richer source of unlabelled drug data to extract more relevant features, which help improving the performance of a predictor learned only from a much smaller set of labelled data. This is one of the key advantages of \mname over previous efforts on DDI prediction~\cite{ryu2018deep}.

\subsection{Interpretable Prediction with Dictionary Learning}
\label{prediction}
Instead of feeding the functional representation through a simple logistic regression model, we develop a dictionary learning module to help human practitioner to understand how \mname makes prediction and identify which sub-structures can possibly lead to the interaction.

\subsubsection{Deep Dictionary Representation.} To provide a more parsimonious representation that also facilitates prediction interpretation, \mname first generates a functional representation $\mathbf{u}_i$ for each frequent sub-structure $\mathbf{C}_i \in \mathbb{C}$ as a single-hot vector $\mathbf{u}_i = [\mathbf{u}_i^{(1)} \ldots \mathbf{u}_i^{(k)}]$ where $\mathbf{u}_i^{(j)} = \mathbb{I}(i = j)$. \mname then exploits the above encoder to generate a matrix $\mathbf{B} = [\mathbf{b}_1, \mathbf{b}_2, \ldots, \mathbf{b}_k]$ of latent feature vectors for $\mathbb{U} = \{\mathbf{u}_1, \ldots, \mathbf{u}_k\}$ such that,
\begin{equation}
\mathbf{b}_i = \mathbf{W}_\mathrm{e}\mathbf{u}_i + \mathbf{b}_\mathrm{e} \ . \label{eq_b}
\end{equation}
Likewise, each functional representation $\mathbf{x}$ of any drug-drug pair $(\mathbf{S},\mathbf{S}')$ (see Definition 2) can also be first translated into a latent feature vector $\mathbf{z}$ using the same encoder (see Eq.~\eqref{eq_encoder}). The resulting latent vector $\mathbf{z}$ can then be projected on a subspace defined by $\mathrm{span}(\mathbf{B})$ for which $\mathbf{z} \simeq \mathbf{b}_1 r_1 + \ldots \mathbf{b}_k r_k$ where $\mathbf{r} = [r_1~r_2~\ldots~r_k]$ is a column vector of projection coefficients. These coefficients can be computed via optimizing the following projection loss,
\begin{equation}
\mathbf{L}_p(\mathbf{r}) = \frac{1}{2}\left\|\mathbf{z} - \mathbf{B}^\top\mathbf{r}\right\|_2^2 + \lambda_1 \|\mathbf{r}\|_2 \label{eq_lasso}
\end{equation}
where the $\mathrm{L}_2$-norm $\|\mathbf{r}\|_2$ controls the complexity of the projection, and $\lambda_1 > 0$ is a regularization parameter. In a more practical implementation, we can also put both the encoder and the above projection (which previously assumes a pre-computed $\mathbf{B}$) in one pipeline to optimize both the basis subspace and the projection. This is achieved via minimizing the following augmented loss function,
\begin{equation}
\mathbf{L}_p\Big(\mathbf{W}_\mathrm{e},\mathbf{b}_\mathrm{e},\mathbf{r}\Big) = \left(\frac{1}{2}\left\|\mathbf{z} - \mathbf{B}^\top\mathbf{r}\right\|_2^2 + \lambda_1 \|\mathbf{r}\|_2\right) + \lambda_2 \left\|\mathbf{B}\right\|_{\mathrm{F}}^2 \ , \label{eq_p}
\end{equation}
where the expression for each column $\mathbf{b}_i$ of $\mathbf{B}$ in Eq.~\eqref{eq_b} is plugged into the above to optimize the encoder parameters $(\mathbf{W}_\mathrm{e},\mathbf{b}_\mathrm{e})$.
Note that the first term in Eq.~\eqref{eq_p} can be minimized more efficiently alone since it can be cast as solving a ridge linear regression task where given a vector $\mathbf{z}$ to be projected and a projection base $\mathbf{B}$, we want to find non-trivial coefficients $\mathbf{r}$ that yields minimum projection loss. By warm-starting the optimization of $\mathbf{r}$ by initializing it with the solution $\mathbf{r}_\ast = \operatorname{\arg\min}_{\mathbf{r}}\mathbf{L}_p(\mathbf{r})$ of the above ridge regression task (see Eq.~\eqref{eq_lasso}), the optimizing process converges faster than randomly initializing $\mathbf{r}$. The above problem can be solved analytically with a closed-form solution:
\begin{equation}
\mathbf{r}_\ast = \Big( \mathbf{B}^\top \mathbf{B} + \lambda_1 \mathbf{I} \Big)^{-1} \mathbf{B}^\top \mathbf{z} \ ,
\label{eq_close}
\end{equation}
where $\mathbf{I}$ is the identity matrix of size $k$. Empirically, we observed that with this warmed-up, the training pipeline can back-propagated through Eq.~\eqref{eq_p} more efficiently. Once optimized, the resulting projection coefficient $\mathbf{r} = [r_1~r_2 \ldots~r_k]$ can be used as a dictionary representation for the corresponding drug-drug pair $(\mathbf{S},\mathbf{S}')$.

\subsubsection{End-to-End Training.} Given each labelled drug-drug pair $(\mathbf{S},\mathbf{S}')$, we generate its functional representation $\mathbf{x} \in [0,1]^k$ (Section~\ref{representation}), latent feature embedding $\mathbf{z} \in \mathbb{R}^d$ and then deep dictionary representation $\mathbf{r}$ as detailed above. A neural network (NN) parameterized by weights $\mathbf{W}_\mathrm{p}$ and biases $\mathbf{b}_\mathrm{p}$ is then constructed to compute a probability score $p(\mathbf{x}) \in [0, 1]$ based on $\mathbf{r}$ to assess the chance that $\mathbf{S}$ and $\mathbf{S}'$ has an interaction. This is defined below:
\begin{equation}
p(\mathbf{x}) = \sigma\Big(\mathbf{W}_\mathrm{p}\mathbf{r} + \mathbf{b}_\mathrm{p}\Big) \ .
\label{eq_prediction}
\end{equation}
We then optimize $\mathbf{W}_\mathrm{p}(\mathbf{x})$ and $\mathbf{b}_\mathrm{p}$ via minimizing the following binary cross-entropy loss with the true interaction outcome $y \in \{0, 1\}$: 
\begin{equation}
\label{eq_pred_loss}
\mathbf{L}_\mathrm{c}(\mathbf{x}, y) = y\log p(\mathbf{x}) + (1 - y)\log\left(1-p(\mathbf{x})\right) \ .
\end{equation}

More specifically, \mname consists of two training stages. In the first stage, we pre-train the auto-encoder and dictionary learning module with our unlabelled drug-drug and drug-food pairs to let the encoder learns the most efficient representation for any chemical structures (e.g., drugs or food constituents) with the combined loss of $\alpha \mathbf{L}_\mathrm{r} + \beta \mathbf{L}_\mathrm{p}$, where $\alpha$ and $\beta$ are manually adjusted to balance between the relative importance of these losses. Then, we fine-tune the entire learning pipeline with the labelled dataset $\mathbb{D}$ for DDI prediction, using the aggregated loss $\alpha \mathbf{L}_\mathrm{r} + \beta \mathbf{L}_\mathrm{p} + \gamma \mathbf{L}_\mathrm{c}$, where $\gamma$ is also a loss adjustment parameter. 

\subsubsection{Interpretable Prediction.} The learned predictor can then be used to make DDI prediction on new drugs and furthermore, the projection coefficients $\mathbf{r} = [r_1~r_2~\ldots~r_k]$ can also be used to assess the relevance of the basis feature vectors $\mathbf{B} = [\mathbf{b}_1, \mathbf{b}_2, \ldots, \mathbf{b}_k]$ to the prediction outcome. As each basis vector $\mathbf{b}_i$ is associated with a frequent molecular sub-graph $\mathbf{C}_i$, its corresponding coefficient $\mathbf{r}_i$ reveals the statistical importance of having the sub-graph $\mathbf{C}_i$ in the molecular graphs of the drug pairs so that they would interact, thus explaining the rationality behind \mname's prediction. 

\section{Experiment}

\label{experiment}
We design experiments to evaluate \mname\footnote{Source code is at https://github.com/kexinhuang12345/CASTER} to answer the following three questions:
\begin{enumerate}[leftmargin=7mm, nosep]
    \item[\textbf{Q1}:] Does \mname provide more accurate DDI prediction than other strong baselines?
    \item [\textbf{Q2}:] Does the unlabelled data improve the DDI prediction in situations with limited labelled data?
    \item [\textbf{Q3}:] Does \mname dictionary module help interpret its predictions?
\end{enumerate}

\begin{table}[t]
\centering
\caption{Dataset Statistics}
\label{tab:stats}
\begin{tabular}{lcc}
\toprule
& Drugbank (DDI) & BIOSNAP  \\ \hline
\# Drugs & 1,850  &  1,322  \\ 
\# Positive DDI Labels  & 221,523  & 41,520       \\ 
\# Negative Labels   & 221,523 & 41,520 \\ 
\bottomrule
\end{tabular}

\vspace{3mm}

\begin{tabular}{lc}
\toprule
     &  Unlabelled\\ \hline
\# Drugs      & 9,675 \\
\# Food Compounds  & 24,738\\
\# Drug-Drug Pairs & 220,000\\
\# Drug-Food Pairs & 220,000\\ 
\bottomrule
\end{tabular}

\end{table}

\begin{table*}[t]
\centering
    \caption{\mname provides more accurate DDI prediction than other strong baselines. First~/~second row of each method corresponds to results reported on BIOSNAP~/~DrugBank (DDI) dataset respectively. }
    \label{DDI_pred_figure}
    \begin{tabular}{llcccc}
    \toprule
    Model & Dataset\ \ \ \ \ & ROC-AUC & PR-AUC & F1 & \# Parameters \\ \hline 
    \multirow{2}{*}{LR} & BIOSNAP & $0.802\pm0.001$ & $0.779 \pm 0.001$& $0.741\pm0.002$ & \multirow{2}{*}{1,723} \\
    & DrugBank & $0.774\pm0.003$ & $0.745\pm0.005$&$0.719\pm0.006$ & \\ \hline
    \multirow{2}{*}{Nat.Prot~\cite{vilar2014similarity}} & BIOSNAP & $0.853\pm0.001$ & $0.848 \pm 0.001$& $0.714\pm0.001$ & \multirow{2}{*}{N/A} \\
    & DrugBank & $0.786\pm 0.003$ & $0.753\pm0.003$&$0.709\pm0.004$ & \\ \hline
    \multirow{2}{*}{Mol2Vec~ \cite{jaeger2018mol2vec}} & BIOSNAP & $0.879\pm0.006$ & $0.861 \pm 0.005$& $0.798\pm0.007$ & \multirow{2}{*}{8,061,953} \\
    & DrugBank & $0.849\pm0.004$ & $0.828 \pm 0.006$& $0.775\pm0.004$ & \\ \hline
    \multirow{2}{*}{MolVAE~ \cite{gomez2018automatic}} & BIOSNAP &  $0.892\pm0.009$ & $0.877 \pm 0.009$& $0.788\pm0.033$ & \multirow{2}{*}{8,012,292} \\
    & DrugBank & $0.852\pm0.006$ & $0.828 \pm 0.009$& $0.769\pm0.031$ & \\ \hline
    \multirow{2}{*}{DeepDDI~\cite{ryu2018deep}} & BIOSNAP & $0.886\pm0.007$ & $0.871\pm 0.007$& $0.817\pm0.007$ & \multirow{2}{*}{8,517,633} \\
    & DrugBank & $0.844\pm0.003$ & $0.828 \pm 0.002$& $0.772\pm0.006$ & \\ \hline
     \multirow{2}{*}{\mname} & BIOSNAP & \bf $0.910 \pm 0.005$ & \bf $0.887 \pm 0.008$ & \bf $0.843 \pm 0.005$ & \multirow{2}{*}{7,813,429} \\
    & DrugBank & \bf $0.861 \pm 0.005$ & \bf $0.829 \pm 0.003$ & \bf $0.796 \pm 0.007$ & \\
    \bottomrule
    \end{tabular}
\end{table*}

\subsection{Experimental Setup}

\subsubsection{Datasets.}
We evaluated \mname using two datasets. (1) {\bf DrugBank (DDI)}~\cite{drugbank} includes 1,850 approved drugs. Each drug is associated with its chemical structure (SMILES), indication, protein binding, weight, mechanism and etc. The data includes 221,523 DDI positive labels. (2) {\bf BIOSNAP}~\cite{biosnapnets} that consists of 1,322 approved drugs with 41,520 labelled DDIs, obtained through drug labels and scientific publications. For both dataset, we generate negative counterparts by sampling complement set of positive drug pairs set as the negative set following standard practice~\cite{zitnik2018modeling}. 

To demonstrate \mname's generalizability to unseen drugs and other chemical compounds, we generate a dataset consisting of unlabelled drug-drug and drug-food compounds pairs. We consider all drugs from DrugBank including experimental, nutraceutical, investigational types of drugs and the food constituent SMILES strings come from FooDB~\cite{foodb}. In total, there are 9,675 drugs and 24,738 food constituents. We randomly generate 220,000 drug-drug pairs and 220,000 drug-food pairs as the unlabelled dataset to be pretrained. Data statistics are summarized in Table~\ref{tab:stats}. For data processing, see (Appendix.~\ref{sec:imple}).

\subsubsection{Metrics.} \label{metrics} We denote $\mathbf{Y}, \widehat{\mathbf{Y}}$ as the ground truth, and predicted values for drug-drug pairs dataset respectively. We measure the prediction accuracy and efficiency using the following metrics:

\begin{enumerate}[leftmargin=*,noitemsep, topsep=0pt]
    \item \textbf{ROC-AUC}: Area under the receiver operating characteristic curve: the area under the plot of the true positive rate against the false positive rate at various thresholds.
    \item \textbf{PR-AUC}: Area under the precision-recall curve: the area under the plot of the precision rate against recall rate at various thresholds.
    \item\textbf{F1 Score}: F1 Score is the harmonic mean of precision and recall. 
    \begin{eqnarray*}
    \mathrm{F1}  =  2 \cdot \frac{\mathrm{P} \cdot \mathrm{R}}{\mathrm{P} + \mathrm{R}},~ \mathrm{where }\ \mathrm{P} =  \frac{\vert \mathbf{Y} \cap \widehat{\mathbf{Y}} \vert}{\vert \mathbf{Y} \vert},  \mathrm{R}  =  \frac{\vert \mathbf{Y} \cap \widehat{\mathbf{Y}} \vert}{\vert \widehat{\mathbf{Y}} \vert}
    \end{eqnarray*}
    \item \textbf{\# Parameters}: The number of parameters used in the model. 
\end{enumerate}

\subsubsection{\mname Setup.} We found the following hyperparameters a best fit to \mname. For SPM, we set the frequency threshold $\eta$ as 50 and it result in $k = 1,722$ frequent substructures. The dimension of latent representation from encoder $d$ is set to be $50$. The encoder and decoder both use three layer perceptrons of hidden size 500 with ReLU activation function. The predictor uses six layer perceptrons with first three layers of hidden size 1,024 and last layer 64. The predictor applies 1-D batch normalization with ReLU activation units. For the trade-off between different loss, we set $\alpha = 1e-1, \beta = 1e-1, \gamma = 1$. For regularization coefficients. we set $\lambda_1 = 1e-5, \lambda_2 = 1e-1$. We first pretrain the network for one epoch with the unlabelled dataset and then proceed to the labelled dataset training process. We use batch size 256 with Adam optimizer of learning rate $1e-3$. All methods are implemented in PyTorch~\cite{pytorch}. For training, we use a server with 2 Intel Xeon E5-2670v2 2.5GHz CPUs, 128 GB RAM and 3 NVIDIA Tesla K80 GPUs.

\subsubsection{Evaluation Strategies.} We randomly divided the dataset into training, validation and testing sets in a 7:1:2 ratio. For each experiment, we conducted $5$ independent runs with different random splits of data and used early stopping based on the \textbf{ROC-AUC} performance on the validation set. The selected model via validation is then evaluated on the test set. The results are reported below.

\subsection{Q1: \mname achieves higher accuracy in DDI prediction}
To answer \textbf{Q1}, we compared  \mname with the following end-to-end algorithms:
\begin{enumerate} [leftmargin=*,noitemsep, topsep=0pt]
    \item \textbf{Logistic Regression (LR)}: LR with L2 regularization using representation generated from SPM.
    \item \textbf{Nat.Prot}: \cite{vilar2014similarity} is the standard approach in real-world DDI prediction task. It uses a similarity-based matrix heuristic method.
    \item \textbf{Mol2Vec}: \cite{jaeger2018mol2vec} applies Word2Vec on ECFP fingerprint of chemical compounds to generate a dense representation of the chemical structures. We adapt it by concatenating the latent variable of input pairs and use large multi-layer perceptron predictors to predict the drugs interaction.
    \item \textbf{MolVAE}: \cite{gomez2018automatic} uses variational autoencoders on SMILES input with the molecular property prediction auxiliary task to generate a compact representation. They use convolutional encoder and GRU-based decoder~\cite{cho2014learning}. We apply this framework by following the same architecture but changing the auxiliary task to drug interaction prediction. 
    \item \textbf{DeepDDI}: ~\cite{ryu2018deep} is a drug interaction prediction model based on a task-specific chemical similarity profile. We modified the last layer of the original DeepDDI implementation from multi-class to a binary class.
\end{enumerate}
DDI prediction performance of all tested methods are reported in Table~\ref{DDI_pred_figure}. The results show that \mname achieves better predictive performance on DDI prediction across all evaluation metrics. This demonstrates that \mname is able to capture the necessary interaction mechanism and is therefore more suitable for drug interaction prediction task than previous approaches. 
\begin{figure*}[t]
    \centering
    
        \includegraphics[width = 0.35\textwidth]{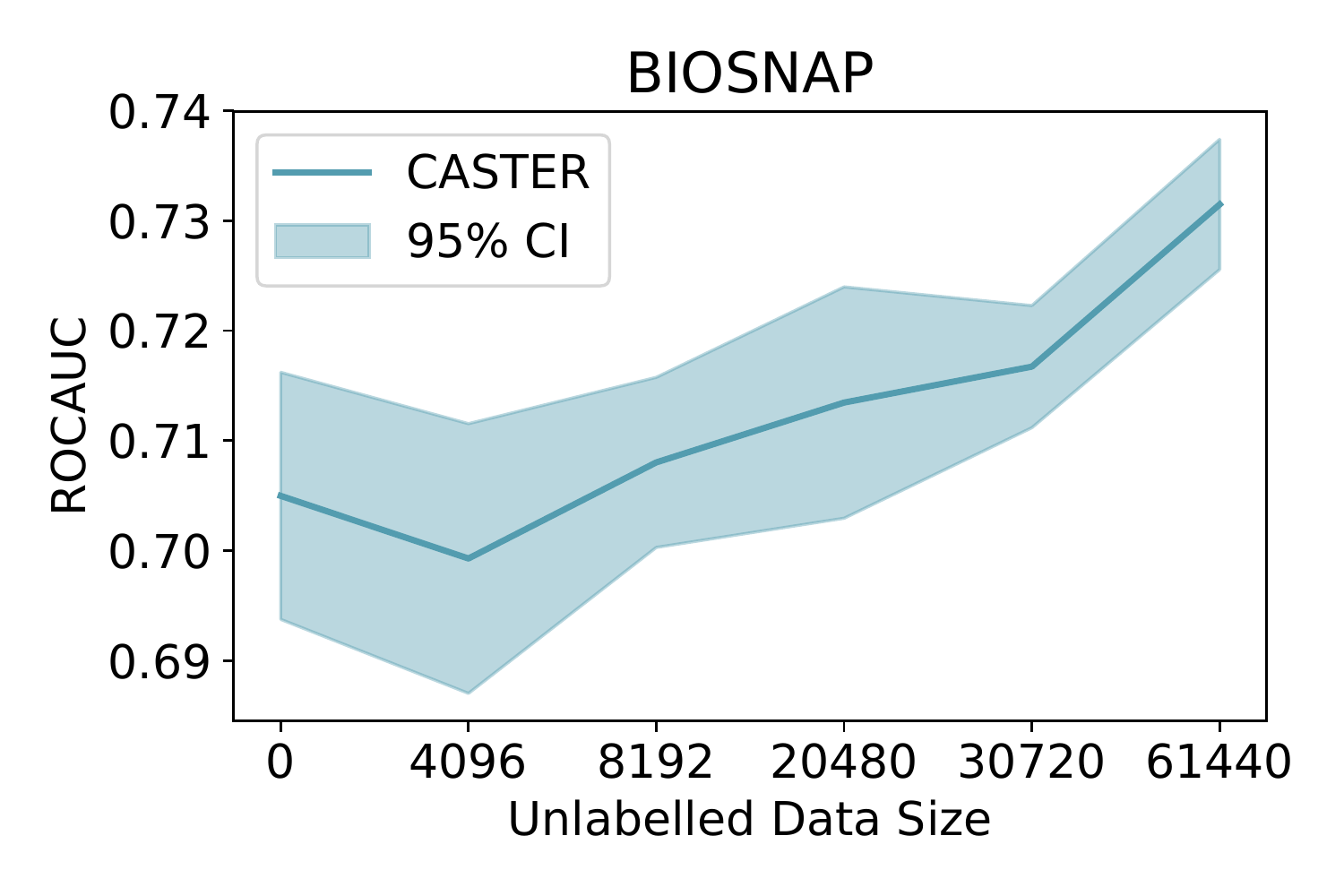}
        \includegraphics[width = 0.35\textwidth]{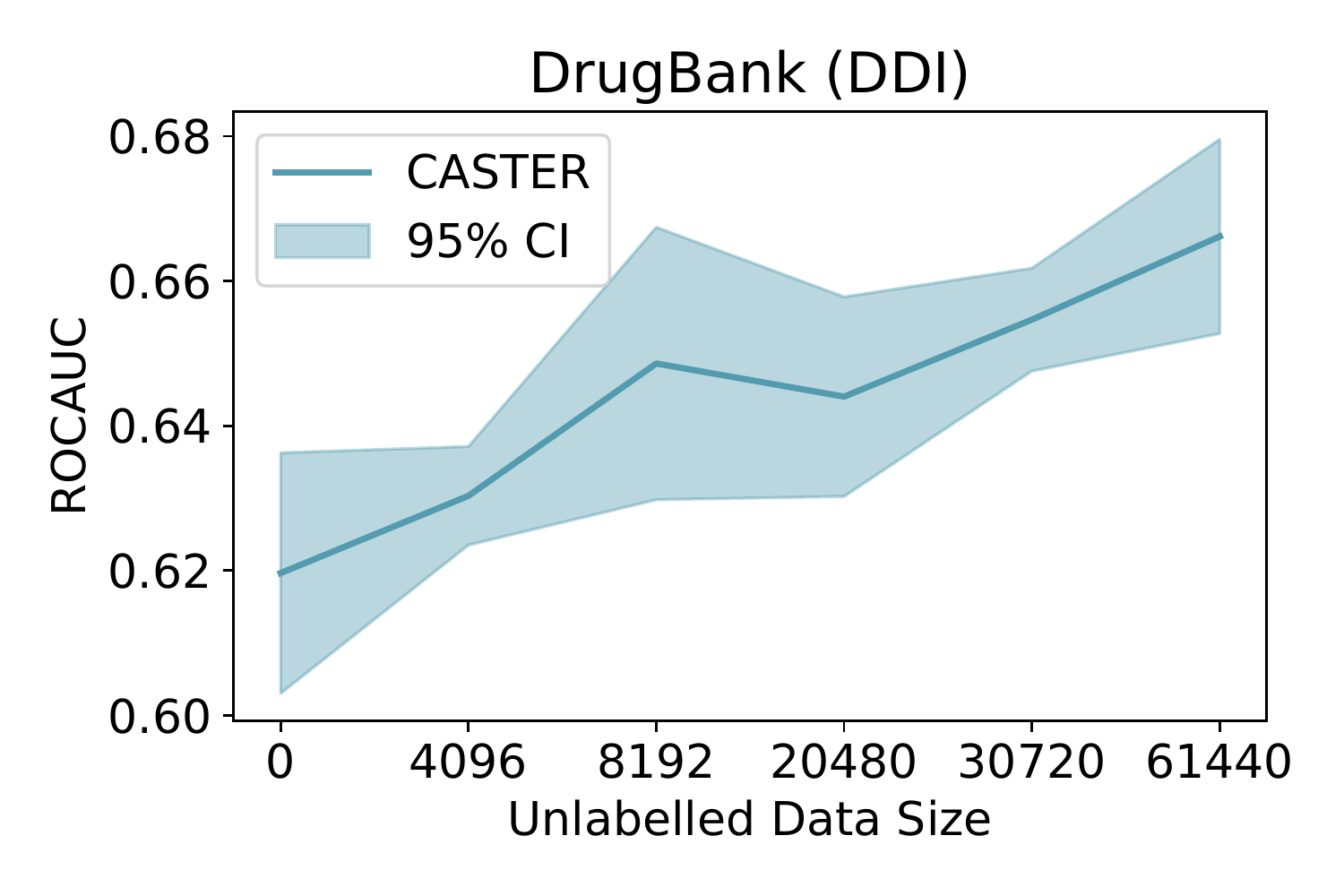}
    \caption{\mname can leverage unlabelled data to improve DDI prediction in scarce labels scenario.}
    
    \label{semi_figure}
\end{figure*}
\begin{figure*}[t]
    \centering
    \includegraphics[width =0.7\textwidth]{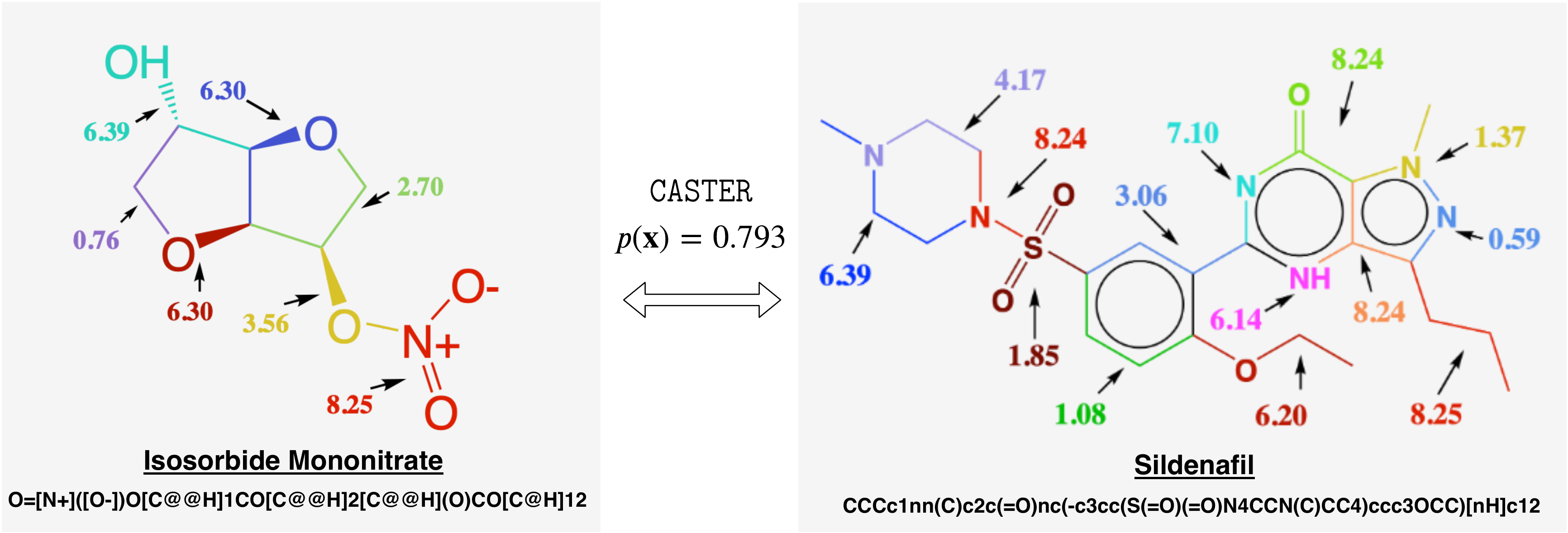}
    \caption{\mname produces interpretable coefficients for understanding  drug interaction mechanism.}
    \label{case}
\end{figure*}

\subsection{Q2: \mname leverages unlabelled data successfully to improve prediction performance}

To answer $\textbf{Q2}$, we design an experiment to use only a small set of labelled data and adjust the unlabelled data size to see if the increasing amount of unlabelled data improves the performance of DDI prediction on the test set. From Fig.~\ref{semi_figure}, we observe that as the amount of unlabelled data increases, \mname is able to leverage more information from the unlabelled data and consistently improves the accuracy of its DDI prediction on both datasets.  
\subsection{Q3: \mname generates interpretable prediction -- A case study}
Given an input pair, \mname generates a set of coefficients showing the relevance of substructures in terms of  \mname's predicted DDIs, a feature missing from other baselines. We select the interaction between Sildenafil and other Nitrate-Based drugs as a case study. 
Sildenafil is an effective treatment for erectile dysfunction and pulmonary hypertension~\cite{langtry1999sildenafil}. Sildenafil is developed as a phosphodiesterase-5 (PDE5) inhibitor. In the presence of PDE5 inhibitor, nitrate ($\textit{NO}_3^-$)-based medications such as Isosorbide Mononitrate (IM) can cause dramatic increase in cyclic guanosine monophosphate~\cite{murad1986cyclic}, which leads to intense drops in blood pressure that can cause heart attack~\cite{langtry1999sildenafil,ishikura2000effects,chamsi2001sildenafil}. To demonstrate \mname's ability to identify potential causes of interaction, we want to test if \mname assigns high coefficient to the nitrate group when it is asked to predict the interaction outcome between Sildenafil and Nitrate-Based drugs. 

We first feed the SMILES strings of Sildenafil and IM into \mname, and it outputs 0.7928 confidence score of interaction, which is very high. We then examine the coefficients corresponding to the functional substructures of the input, and observe that nitrate (\textit{O=[N+]([O-])}) in IM has the highest coefficient\footnote{Note that this is the augmented coefficients multiplied by a factor, see~Appendix~\ref{sec:imple}} of 8.25 among 21 functional substructures identified by \mname, which matches the chemical intuition discussed above. Fig.~\ref{case} illustrates the coefficients generated by \mname for different sub-structures. To test the robustness of the interpretability result, we show that random initialization does not affect our prediction by training five models with different random seeds and calculated the coefficients in Fig.~\ref{case}. We consider coefficient from each run as a variable and then calculated their Correlation Matrix. The high average correlation (\textbf{0.7673}) implies \mname is robust based on \cite{mukaka2012guide}. In order to avoid if the result is due to chance, we search for all nitrate-based drugs in BIOSNAP dataset, which results in the other four drugs: Nitroglycerin, Isosorbide Dinitrate, Silver Nitrate, and Erythrityl Tetranitrate. We pair the SMILES string of Sildenafil with each of these nitrate drugs and feed them to \mname. We then observe that \mname assigns on average 50\% higher coefficient to nitrate than the mean of coefficients of other sub-structures existed in the input pair, which is not a coincidence. This case study thus demonstrated that \mname is capable of suggesting sparse and reasonable cues on finding substructures which are likely responsible to DDI.

\section{Conclusion}

This paper developed a new computational framework for DDI prediction called \mname, which is an end-to-end dictionary learning framework that incorporates a specialized representation for DDI prediction, inspired by the chemical mechanism of drug interactions. We demonstrated empirically that \mname can provide more accurate and interpretable DDI predictions than the previous approaches that use generic drug representations. For future works, we plan to extend it to chemical sub-graph embedding and incorporate metric learning for further improvement.

\section*{Acknowledgement}
This work was in part supported by the National Science Foundation award IIS-1418511, CCF-1533768 and IIS-1838042, the National Institute of Health award NIH R01 1R01NS107291-01 and R56HL138415.

\appendix
\section{Implementation Details}\label{sec:imple}

For preprocessing of the SMILES string, we use the RDKit software to convert all string into canonical form. As the software first maps the string into a Mol format file, there are cases where RDKit cannot process some SMILES strings. We omit these data points. The unique number of drugs decrease from 2,159 to 1,850 for DrugBank dataset and 1,514 to 1,322 for BIOSNAP. The overall number of DDI positive samples drop from 222,127 to 221,523 for DrugBank and 48,514 to 41,520 for BIOSNAP.

Due to the large size of DrugBank and limited time, experiment done on the DrugBank dataset uses another dataset split scheme to not only decrease the time but also fully leverage the dataset. For the five independent run in each experiment, we divide the whole set into exclusive five folds data. We then report the average and standard deviation scores across these five runs. Hence, the evaluation considers the full dataset but only uses a subsampled dataset in each independent run. 

\noindent\textbf{Magnifying Factor.}\label{mag} During training \mname, we observe that since Eq.~\eqref{eq_close} applies ridge regression, the coefficients $\mathbf{r}$ are relatively small. In order to allow faster training for the predictor, we multiple each coefficient by a magnifying factor of 100, i.e. enlarging the learning rate for the input weights, to obtain an augmented coefficient. This allow the loss decreases much faster.
\bibliography{main}

\begin{thebibliography}{}

\bibitem[\protect\citeauthoryear{Chamsi-Pasha}{2001}]{chamsi2001sildenafil}
Chamsi-Pasha, H.
\newblock 2001.
\newblock Sildenafil (viagra) and the heart.
\newblock {\em Journal of family \& community medicine} 8(2):63.

\bibitem[\protect\citeauthoryear{Cho \bgroup et al\mbox.\egroup
  }{2014}]{cho2014learning}
Cho, K.; van Merri{\"e}nboer, B.; Gulcehre, C.; Bahdanau, D.; Bougares, F.;
  Schwenk, H.; and Bengio, Y.
\newblock 2014.
\newblock Learning phrase representations using {RNN} encoder{--}decoder for
  statistical machine translation.
\newblock In {\em Proceedings of the 2014 Conference on Empirical Methods in
  Natural Language Processing ({EMNLP})},  1724--1734.

\bibitem[\protect\citeauthoryear{Ferdousi, Safdari, and
  Omidi}{2017}]{Ferdousi17}
Ferdousi, R.; Safdari, R.; and Omidi, Y.
\newblock 2017.
\newblock Computational prediction of drug-drug interactions based on drugs
  functional similarities.
\newblock {\em Journal of Biomedical Information}.

\bibitem[\protect\citeauthoryear{FooDB}{2018}]{foodb}
FooDB.
\newblock 2018.
\newblock Foodb version 1.0.
\newblock \url{http://foodb.ca/}.

\bibitem[\protect\citeauthoryear{Fu, Xiao, and Sun}{2020}]{core}
Fu, T.; Xiao, C.; and Sun, J.
\newblock 2020.
\newblock Core: Automatic molecule optimization using copy \& refine strategy.
\newblock {\em AAAI Conference on Artificial Intelligence}.

\bibitem[\protect\citeauthoryear{Gage}{1994}]{gage1994new}
Gage, P.
\newblock 1994.
\newblock A new algorithm for data compression.
\newblock {\em The C Users Journal} 12(2):23--38.

\bibitem[\protect\citeauthoryear{Giacomini \bgroup et al\mbox.\egroup
  }{2007}]{Giacomini2007}
Giacomini, K.; Krauss, R.; Roden, D.; Eichelbaum, M.; and Hayden, M.
\newblock 2007.
\newblock When good drugs go bad.
\newblock {\em Nature} 446:975--977.

\bibitem[\protect\citeauthoryear{G{\'o}mez-Bombarelli \bgroup et
  al\mbox.\egroup }{2018}]{gomez2018automatic}
G{\'o}mez-Bombarelli, R.; Wei, J.~N.; Duvenaud, D.; Hern{\'a}ndez-Lobato,
  J.~M.; S{\'a}nchez-Lengeling, B.; Sheberla, D.; Aguilera-Iparraguirre, J.;
  Hirzel, T.~D.; Adams, R.~P.; and Aspuru-Guzik, A.
\newblock 2018.
\newblock Automatic chemical design using a data-driven continuous
  representation of molecules.
\newblock {\em ACS central science} 4(2):268--276.

\bibitem[\protect\citeauthoryear{Huang \bgroup et al\mbox.\egroup
  }{2019}]{espf}
Huang, K.; Xiao, C.; Glass, L.; and Sun, J.
\newblock 2019.
\newblock Explainable substructure partition fingerprint for protein, drug, and
  more.
\newblock {\em NeurIPS Learning Meaningful Representation of Life Workshop}.

\bibitem[\protect\citeauthoryear{Ishikura \bgroup et al\mbox.\egroup
  }{2000}]{ishikura2000effects}
Ishikura, F.; Beppu, S.; Hamada, T.; Khandheria, B.~K.; Seward, J.~B.; and
  Nehra, A.
\newblock 2000.
\newblock Effects of sildenafil citrate (viagra) combined with nitrate on the
  heart.
\newblock {\em Circulation} 102(20):2516--2521.

\bibitem[\protect\citeauthoryear{Jaeger, Fulle, and
  Turk}{2018}]{jaeger2018mol2vec}
Jaeger, S.; Fulle, S.; and Turk, S.
\newblock 2018.
\newblock Mol2vec: unsupervised machine learning approach with chemical
  intuition.
\newblock {\em Journal of chemical information and modeling} 58(1):27--35.

\bibitem[\protect\citeauthoryear{James}{2004}]{james2004daylight}
James, C.~A.
\newblock 2004.
\newblock Daylight theory manual.

\bibitem[\protect\citeauthoryear{Jin \bgroup et al\mbox.\egroup
  }{2017}]{AAAI1714292}
Jin, B.; Yang, H.; Xiao, C.; Zhang, P.; Wei, X.; and Wang, F.
\newblock 2017.
\newblock Multitask dyadic prediction and its application in prediction of
  adverse drug-drug interaction.

\bibitem[\protect\citeauthoryear{Langtry and
  Markham}{1999}]{langtry1999sildenafil}
Langtry, H.~D., and Markham, A.
\newblock 1999.
\newblock Sildenafil.
\newblock {\em Drugs} 57(6):967--989.

\bibitem[\protect\citeauthoryear{Lo \bgroup et al\mbox.\egroup
  }{2018}]{lo2018machine}
Lo, Y.-C.; Rensi, S.~E.; Torng, W.; and Altman, R.~B.
\newblock 2018.
\newblock Machine learning in chemoinformatics and drug discovery.
\newblock {\em Drug discovery today}.

\bibitem[\protect\citeauthoryear{Ma \bgroup et al\mbox.\egroup
  }{2018}]{ma2018drug}
Ma, T.; Xiao, C.; Zhou, J.; and Wang, F.
\newblock 2018.
\newblock Drug similarity integration through attentive multi-view graph
  auto-encoders.
\newblock In {\em Proceedings of the Twenty-Seventh International Joint
  Conference on Artificial Intelligence, {IJCAI-18}},  3477--3483.
\newblock International Joint Conferences on Artificial Intelligence
  Organization.

\bibitem[\protect\citeauthoryear{Marinka~Zitnik and
  Leskovec}{2018}]{biosnapnets}
Marinka~Zitnik, Rok~Sosi\v{c}, S.~M., and Leskovec, J.
\newblock 2018.
\newblock {BioSNAP Datasets}: {Stanford} biomedical network dataset collection.
\newblock \url{http://snap.stanford.edu/biodata}.

\bibitem[\protect\citeauthoryear{Mukaka}{2012}]{mukaka2012guide}
Mukaka, M.~M.
\newblock 2012.
\newblock A guide to appropriate use of correlation coefficient in medical
  research.
\newblock {\em Malawi Medical Journal} 24(3):69--71.

\bibitem[\protect\citeauthoryear{Murad}{1986}]{murad1986cyclic}
Murad, F.
\newblock 1986.
\newblock Cyclic guanosine monophosphate as a mediator of vasodilation.
\newblock {\em The Journal of clinical investigation} 78(1):1--5.

\bibitem[\protect\citeauthoryear{Onakpoya, Heneghan, and
  Aronson}{2016}]{onakpoya2016post}
Onakpoya, I.~J.; Heneghan, C.~J.; and Aronson, J.~K.
\newblock 2016.
\newblock Post-marketing withdrawal of 462 medicinal products because of
  adverse drug reactions: a systematic review of the world literature.
\newblock {\em BMC medicine} 14(1):10.

\bibitem[\protect\citeauthoryear{Paszke \bgroup et al\mbox.\egroup
  }{2017}]{pytorch}
Paszke, A.; Gross, S.; Chintala, S.; Chanan, G.; Yang, E.; DeVito, Z.; Lin, Z.;
  Desmaison, A.; Antiga, L.; and Lerer, A.
\newblock 2017.
\newblock Automatic differentiation in pytorch.

\bibitem[\protect\citeauthoryear{Riniker and Landrum}{2013}]{riniker2013open}
Riniker, S., and Landrum, G.~A.
\newblock 2013.
\newblock Open-source platform to benchmark fingerprints for ligand-based
  virtual screening.
\newblock {\em Journal of cheminformatics} 5(1):26.

\bibitem[\protect\citeauthoryear{Rogers and Hahn}{2010}]{rogers2010extended}
Rogers, D., and Hahn, M.
\newblock 2010.
\newblock Extended-connectivity fingerprints.
\newblock {\em Journal of chemical information and modeling} 50(5):742--754.

\bibitem[\protect\citeauthoryear{Ryu, Kim, and Lee}{2018}]{ryu2018deep}
Ryu, J.~Y.; Kim, H.~U.; and Lee, S.~Y.
\newblock 2018.
\newblock Deep learning improves prediction of drug--drug and drug--food
  interactions.
\newblock {\em Proceedings of the National Academy of Sciences}
  115(18):E4304--E4311.

\bibitem[\protect\citeauthoryear{Sennrich, Haddow, and
  Birch}{2016}]{sennrich2015neural}
Sennrich, R.; Haddow, B.; and Birch, A.
\newblock 2016.
\newblock Neural machine translation of rare words with subword units.
\newblock In {\em Proceedings of the 54th Annual Meeting of the Association for
  Computational Linguistics},  1715--1725.
\newblock Berlin, Germany: Association for Computational Linguistics.

\bibitem[\protect\citeauthoryear{Silverman and
  Holladay}{2014}]{silverman2014organic}
Silverman, R.~B., and Holladay, M.~W.
\newblock 2014.
\newblock {\em The organic chemistry of drug design and drug action}.
\newblock Academic press.

\bibitem[\protect\citeauthoryear{System}{2015}]{smiles}
System, D. C.~I.
\newblock 2015.
\newblock Smiles tutorial.

\bibitem[\protect\citeauthoryear{Vilar \bgroup et al\mbox.\egroup
  }{2014}]{vilar2014similarity}
Vilar, S.; Uriarte, E.; Santana, L.; Lorberbaum, T.; Hripcsak, G.; Friedman,
  C.; and Tatonetti, N.~P.
\newblock 2014.
\newblock Similarity-based modeling in large-scale prediction of drug-drug
  interactions.
\newblock {\em Nature protocols} 9(9):2147.

\bibitem[\protect\citeauthoryear{Whitebread \bgroup et al\mbox.\egroup
  }{2005}]{whitehead}
Whitebread, S.; Hamon, J.; Bojanic, D.; and Urban, L.
\newblock 2005.
\newblock In vitro safety pharmacology profiling: an essential tool for
  successful drug development.
\newblock {\em Drug Discovery Today} 10(21).

\bibitem[\protect\citeauthoryear{Wishart \bgroup et al\mbox.\egroup
  }{2008}]{drugbank}
Wishart, D.~S.; Knox, C.; Guo, A.; Cheng, D.; Shrivastava, S.; Tzur, D.;
  Gautam, B.; and Hassanali, M.
\newblock 2008.
\newblock Drugbank: a knowledgebase for drugs, drug actions and drug targets.
\newblock {\em Nucleic Acids Research} 36:901--906.

\bibitem[\protect\citeauthoryear{Xiao \bgroup et al\mbox.\egroup
  }{2017}]{Xiao:2017:ADR:3298239.3298470}
Xiao, C.; Zhang, P.; Chaowalitwongse, W.~A.; Hu, J.; and Wang, F.
\newblock 2017.
\newblock Adverse drug reaction prediction with symbolic latent dirichlet
  allocation.
\newblock In {\em Proceedings of the Thirty-First AAAI Conference on Artificial
  Intelligence}, AAAI'17,  1590--1596.
\newblock AAAI Press.

\bibitem[\protect\citeauthoryear{Zhang \bgroup et al\mbox.\egroup
  }{2015}]{zhang2015label}
Zhang, P.; Wang, F.; Hu, J.; and Sorrentino, R.
\newblock 2015.
\newblock Label propagation prediction of drug-drug interactions based on
  clinical side effects.
\newblock {\em Scientific reports} 5.

\bibitem[\protect\citeauthoryear{Zitnik, Agrawal, and
  Leskovec}{2018}]{zitnik2018modeling}
Zitnik, M.; Agrawal, M.; and Leskovec, J.
\newblock 2018.
\newblock Modeling polypharmacy side effects with graph convolutional networks.
\newblock {\em Bioinformatics} 34(13):i457--i466.

\end{thebibliography}
\bibliographystyle{aaai}

\end{document}